\documentclass[11pt,a4paper]{article}
\usepackage[hyperref]{acl2019}
\usepackage{times}
\usepackage{latexsym}
\usepackage{enumerate}
\usepackage{enumitem}
\usepackage{setspace}
\usepackage{caption}
\usepackage{booktabs}
\usepackage{adjustbox}
\usepackage{xspace}
\usepackage{array,booktabs}
\usepackage{subcaption}

\usepackage{graphicx,calc}
\usepackage{xspace}

\newcommand*{\img}[1]{%
    \raisebox{-.2\baselineskip}{%
        \includegraphics[
        height=1cm,
        width=0.70cm,
        keepaspectratio,
        ]{#1}%
    }%
}

\newcommand*{\bimg}[1]{%
    \raisebox{-.2\baselineskip}{%
        \includegraphics[
        height=.5cm,
        width=.5cm,
        keepaspectratio,
        ]{#1}%
    }%
}

\newcommand*\rh{\img{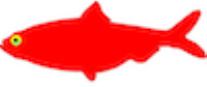}}
\newcommand*\btq{\bimg{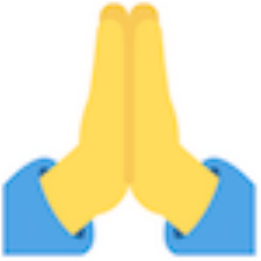}}
\newcommand*\qc{\bimg{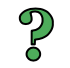}}
\newcommand*\hg{\bimg{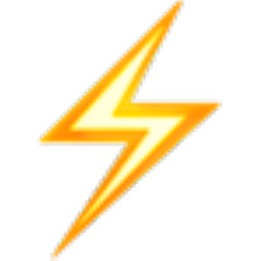}}

\usepackage{url}

\aclfinalcopy 

\setkeys{Gin}{width=\linewidth,totalheight=\textheight,keepaspectratio}

\newcommand{\claim}{\textit{claim}\xspace}
\newcommand{\premise}{\textit{premise}\xspace}

\title{Riposte! A Large Corpus of Counter-Arguments}

\author{
Paul Reisert${}^{\,\dagger,\ddagger}$
\quad\hspace{-5pt}
Benjamin Heinzerling${}^{\,\dagger,\ddagger}$
\quad\hspace{-5pt}
Naoya Inoue${}^{\,\ddagger,\dagger}$
\quad\hspace{-5pt}
Shun Kiyono${}^{\dagger,\ddagger}$
\quad\hspace{-5pt}
Kentaro Inui${}^{\,\ddagger, \dagger}$
\quad\hspace{-5pt}
\\[5pt]
${}^{\dagger}$ RIKEN Center for Advanced Intelligence Project \quad
${}^{\ddagger}$ Tohoku University\\
{\tt \{\hspace{0em}paul.reisert,\hspace{0em}benjamin.heinzerling,\hspace{0em}shun.kiyono\}@riken.jp}\\
{\tt \{\hspace{0em}naoya-i,\hspace{0em}inui\}@ecei.tohoku.ac.jp}
}

\date{}

\begin{document}
\maketitle
\begin{abstract}
Constructive feedback is an effective method for improving critical thinking skills. Counter-arguments (CAs), one form of constructive feedback, have been proven to be useful for critical thinking skills. However, little work has been done for constructing a large-scale corpus of them which can drive research on automatic generation of CAs for fallacious micro-level arguments (i.e.\ a single \claim and \premise pair). In this work, we cast providing constructive feedback as a natural language processing task and create \textbf{Riposte!}, a corpus of CAs, towards this goal. Produced by crowdworkers, \textbf{Riposte!} contains over 18k CAs. We instruct workers to first identify common fallacy types and produce a CA which identifies the fallacy. We analyze how workers create CAs and construct a baseline model based on our analysis. 

\end{abstract}

\section{Introduction}
\label{sec:introduction}

Critical thinking is a crucial skill necessary for valid reasoning, especially for students in a pedagogical context. Towards improving critical thinking skills for students, educators have evaluated the contents of a work and provided constructive feedback (i.e.\ criticism) to the student. Although such methods are effective, they require educators to articulately evaluate the contents of an essay, which can be time-consuming and varies depending on an educator's critical thinking skills.

\begin{figure}[t]
\begin{center}
\includegraphics[clip, width=.5\textwidth]{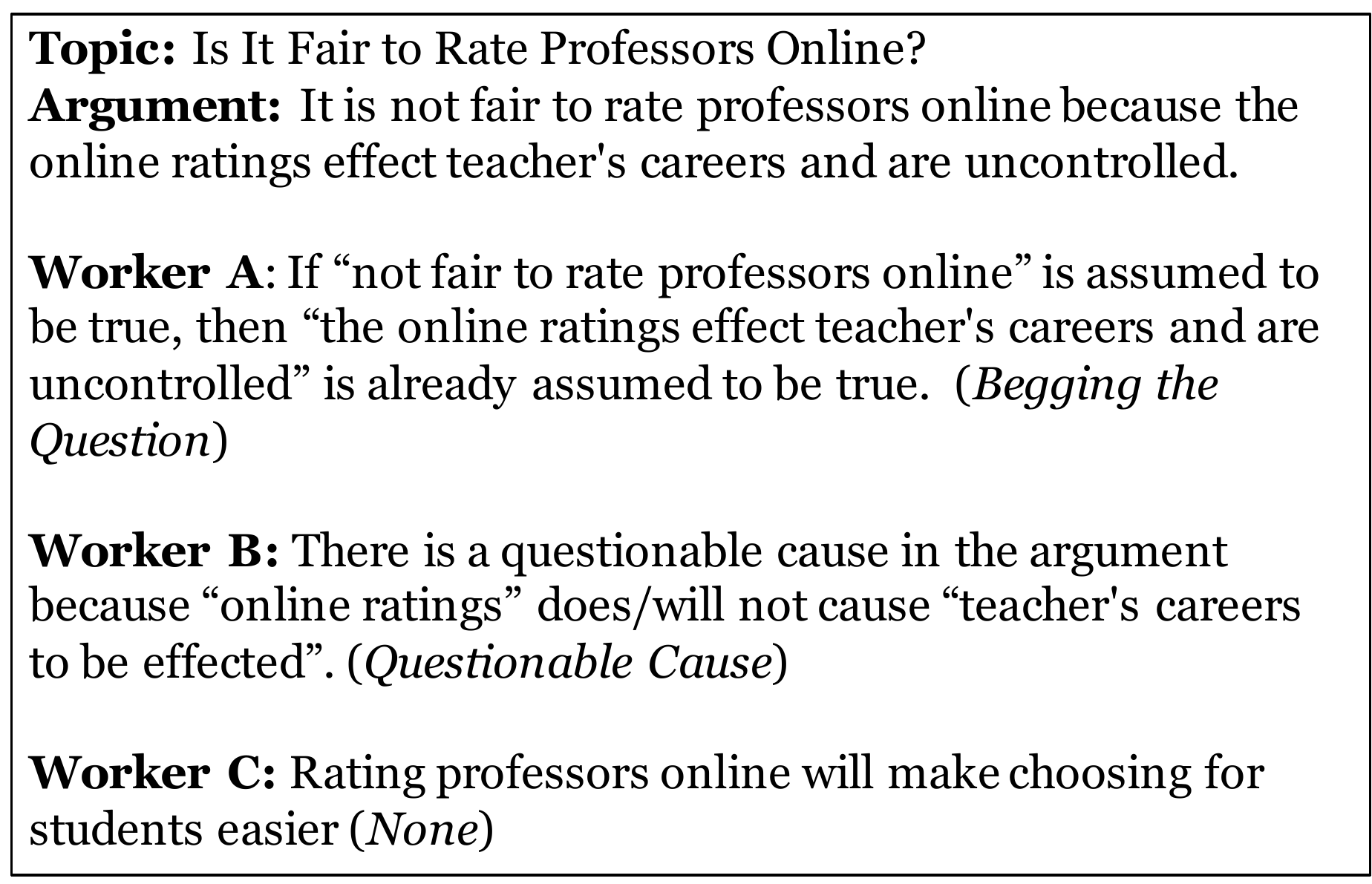}
\caption{CAs in Riposte!\@ produced by crowdworkers. The fallacy type selected by a worker is shown in parentheses.}
\label{fig:corpus}
\end{center}
\end{figure}


\begin{table*}[t!]
    \centering
    \small
    {\renewcommand{\arraystretch}{.1} 
    \begin{tabular}{p{3.65cm}p{5.3cm}p{6cm}}
         \toprule
         \textbf{Fallacy Type} & \textbf{Definition} & \textbf{Template}\\
         \midrule
         \textbf{Begging the Question (}\btq\textbf{)} &
         The truth of the premise is already assumed by the claim.
         &``\textit{If [something] is assumed to be true, then [something else] is already assumed to be true}''.\\
         \textbf{Hasty Generalization(}\hg\textbf{)} &
         Someone assumes something is generally always the case based on a few instances. &
         ``\textit{It's too hasty to assume that [text]}''. \\
         \textbf{Questionable Cause (}\qc\textbf{)} &
         The cause of an effect is questionable.&
         ``\textit{There is a questionable cause in the argument because [questionable cause] does/will not cause [effect]}''.\\
         \textbf{Red Herring (}\rh\textbf{)}&
         Someone reverts attention away from the original claim by changing the topic.&
         ``\textit{The topic being discussed is [first topic], but it is being changed to [second topic]}''.\\
         \bottomrule
    \end{tabular}
    }
    \caption{Definition and templates of fallacy types used in our experiments.}
    \label{tab:types}
\end{table*}

In the field of educational research, the usefulness of identifying fallacies and counter-arguments, henceforth \textit{CAs}, as constructive feedback has been emphasized~\cite{de2008constructive,oktavia2014analysis, indah2015fallacies, song2013teaching}, as both can help writers produce high-quality arguments while simultaneously improving their critical thinking skills. Shown in Figure~\ref{fig:corpus} is an example of an argument with a fallacy  (i.e.\ errors in the logical reasoning of the \textit{argument}) and its CAs (i.e.\ attacks to the \textit{argument}).  In the field of NLP, previous works have addressed fallacy identification~\cite{HABERNAL18.494}, CA retrieval~\cite{wachsmuth2017computational}, and CA generation for macro-level arguments~\cite{hua2018neural}, and essay criteria such as thesis clarity~\cite{persing2013modeling}, argument strength~\cite{persing2015modeling}, and stance~\cite{persing2016modeling} have been evaluated. However, in the pedagogical context, macro-level arguments (e.g.,\ an essay) may consist of several micro-level arguments (i.e.\ one claim/premise pair) that can each contain multiple fallacies. To bridge this gap, we create CAs for micro-level arguments which can be useful for automatic constructive feedback generation.

Several challenges exist for creating a corpus of CAs for constructive feedback. First, the corpus must contain a variety of different topics and arguments to both train and evaluate a model for unseen topics. Second, an argument can have many different fallacies which are not easily identifiable~\cite{oktavia2014analysis, indah2015fallacies, el2017logical}. Third, producing CAs is costly and time-consuming. 

In this work, we design a task for automatic constructive feedback and create \textbf{Riposte!}, a large-scale corpus of CAs via crowdsourcing. Workers are first instructed to identify common fallacy types (\textit{begging the question}, \textit{hasty generalization}, \textit{questionable cause}, and \textit{red herring}) in educational research~\cite{de2008constructive,oktavia2014analysis, indah2015fallacies, song2013teaching} and create a CA for micro-level arguments. In total, we collect 18,887 CAs (see Figure~\ref{fig:corpus} for examples of CAs in Riposte!). We then cast automatic constructive feedback as a text generation task and create a baseline model. 

\begin{table}[t!]
    \centering
    \footnotesize
    \begin{adjustbox}{max width=\linewidth}
    \begin{tabular}{l|r|r|r|r|r}
         \toprule
         \textbf{Criteria} & \btq & \hg & \qc & \rh & \textbf{Total}\\
         \midrule
        Unsure & 2,043 & 315 & 2,136 & 1,879 & 6,373\\\midrule
         CAs (FS) & 3,365 & 3,818 & 2,121 & 1,772 & \textbf{11,076}\\
         CAs (O) & 907 & 2,182 & 2,058 & 2,664 & \textbf{7,811}\\
         \midrule
         \textbf{CAs (total)} & \textbf{4,272} & \textbf{6,000} & \textbf{4,179} & \textbf{4,436} & \textbf{18,887}\\
         \bottomrule
    \end{tabular}
    \end{adjustbox}
    \caption{Full statistics of the Riposte!\@ corpus, where FS represents fallacy-specific CAs and O represents other.}
    \label{tab:distributions}
\end{table}

\section{The Riposte!\@ corpus}
In this section, we determine if training data can easily be created. To the best of our knowledge, this is the first research that addresses corpus construction for automatic constructive feedback.

\subsection{Counter-arguments as an NLP task}

When designing a task for automatic constructive feedback, one must take into account real-world situations. In the pedagogical context, educators can choose the same topic for students annually. With automatic constructive feedback, educators may choose to use a pretrained, supervised model for a single topic with editable background knowledge (i.e., educators can choose which knowledge is necessary to automatically construct feedback). On the other hand, educators may choose a new topic each year, and thus a conditioned model for multiple topics may also be considered. The input to a model should be a \textbf{topic} and several \textbf{claim} and \textbf{premise} argument pairs, and the output would be a set of \textbf{CAs} useful for improving the argument.

\subsection{Existing corpus of arguments}
When training a model for constructive feedback, the data should consist of many CAs for a wide variety of topics. We use the Argument Reasoning Comprehension (ARC) dataset~\cite{Habernal.et.al.2018.NAACL.ARCT}, a corpus of 1,263 unique topic-claim-premise pairs (172 unique topics and 264 unique claims). We assume the arguments in ARC contain many fallacies because they were created by non-expert crowdworkers (i.e., workers are not experts in the field of argumentation).       

\subsection{Riposte!\@ creation}

For creating Riposte!, we use the crowdsourcing platform Amazon Mechanical Turk.\footnote{\url{https://www.mturk.com/}}



\paragraph{\textbf{Data Collection}}
One challenge for collecting training data for automatic constructive feedback is that the CAs should be useful for improving an argument. To assist with collecting such CAs, we adopt \newcite{reisert2019annotation}'s protocol for collecting CAs using crowdsourcing.
We first make several modifications for our data collection (see Appendix). We create 4 separate crowdsourcing tasks (i.e., one for each fallacy type). For each of the 1,263 arguments in ARC, we ask 5 workers to produce a CA. For each fallacy type, we assist workers by providing them with a ``fill-in-the-blank'' template, where workers were instructed to fill in text boxes for a given pattern. The fallacy types and templates are shown in Table~\ref{tab:types}.


\subsection{Riposte!\@ statistics}
The statistics of Riposte!\@ are shown in Table~\ref{tab:distributions}.
11,076 of the CAs are fallacy-specific (i.e.\ workers first identified a fallacy and then created the CA), and 7,811 CAs were created when a worker did not believe the specified fallacy existed in the argument. 6,373 instances were labeled as \textit{unsure} (i.e.\ the worker was unsure about the fallacy type).


\section{How did workers create CAs?}
\label{corpus_eval}



\begin{table}[t!]
    \centering
    \begin{adjustbox}{max width=\linewidth}
    \begin{tabular}{l|r|r|r|r|r}
    \toprule
    \textbf{Criteria} & \btq & \hg & \qc & \rh & \textbf{Total}\\
    \midrule
    Score & 0.61 & 0.17 & 0.35 & 0.36 & 0.24\\
    \bottomrule
    \end{tabular}
    \end{adjustbox}
    \caption{The average Jaccard's similarity scores between CAs for a single argument for each fallacy type.}
    \label{tab:diversity}
\end{table}



\label{ss:similarity}


When creating training data for automatic constructive feedback, CAs should be useful and diverse. We determine how workers create CAs by calculating the similarity between i) a CA and argument and ii) CAs for single arguments.

\paragraph{\textbf{How similar is one CA to the premise-claim?}}

\begin{figure*}[t]
\centering
    \begin{subfigure}[t]{0.4\columnwidth}
        \centering
        \includegraphics[width=\columnwidth, keepaspectratio]
        {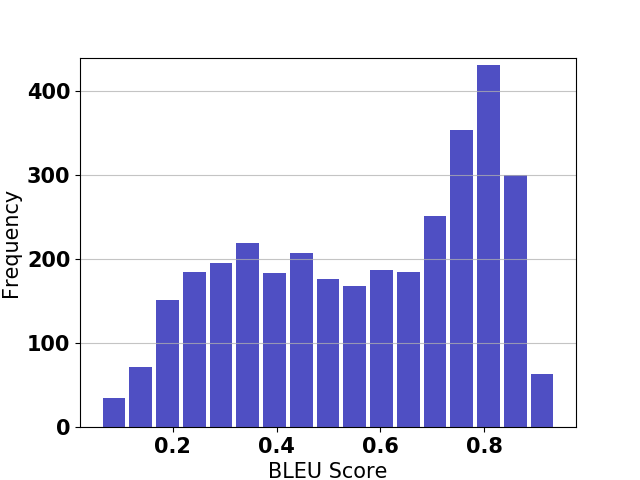}
        \caption{\btq}\label{fig:bleubtq}      
    \end{subfigure}
\centering
    \begin{subfigure}[t]{0.4\columnwidth}
        \centering
        \includegraphics[width=\columnwidth, keepaspectratio]
        {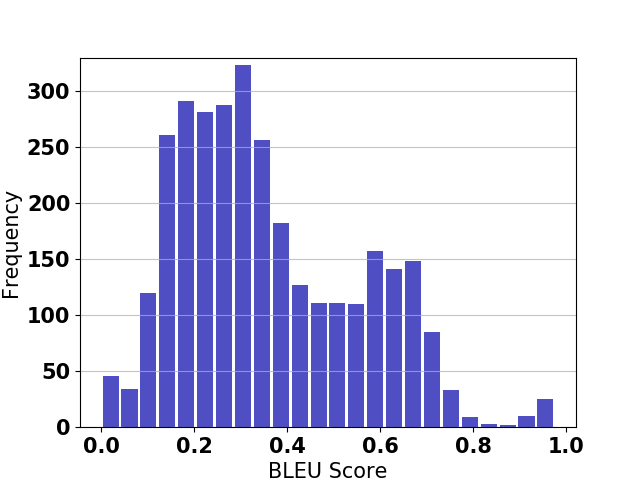}
        \caption{\hg}\label{fig:bleuhg}      
    \end{subfigure}
\centering
    \begin{subfigure}[t]{0.4\columnwidth}
        \centering
        \includegraphics[width=\columnwidth, keepaspectratio]
        {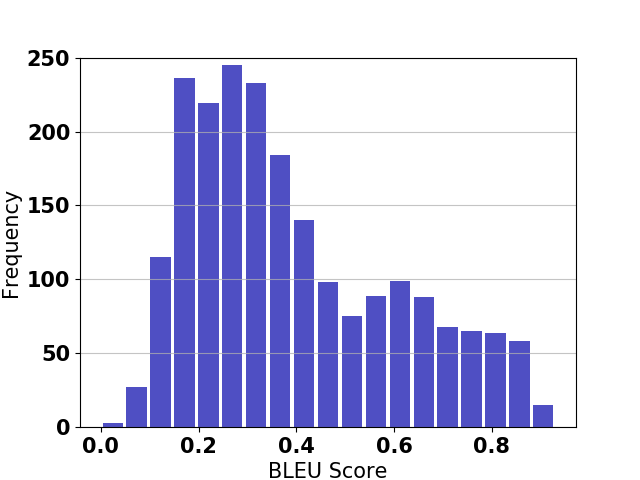}
        \caption{\qc}\label{fig:bleuqc}      
    \end{subfigure}
\centering
    \begin{subfigure}[t]{0.4\columnwidth}
        \centering
        \includegraphics[width=\columnwidth, keepaspectratio]
        {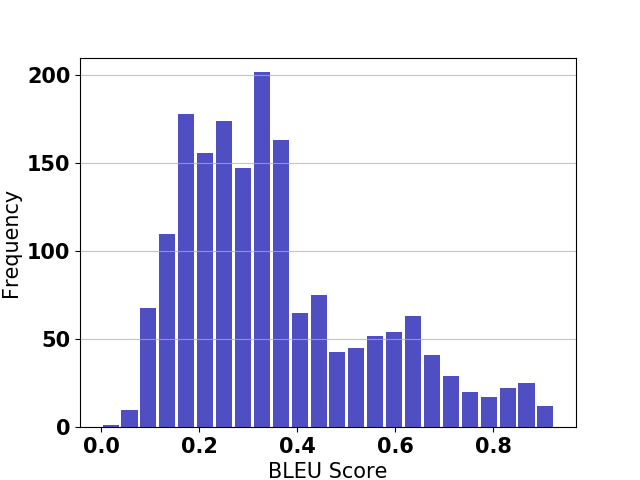}
        \caption{\rh}\label{fig:bleurh}      
    \end{subfigure}
\centering
    \begin{subfigure}[t]{0.4\columnwidth}
    \centering
    \includegraphics[clip, width=\columnwidth]{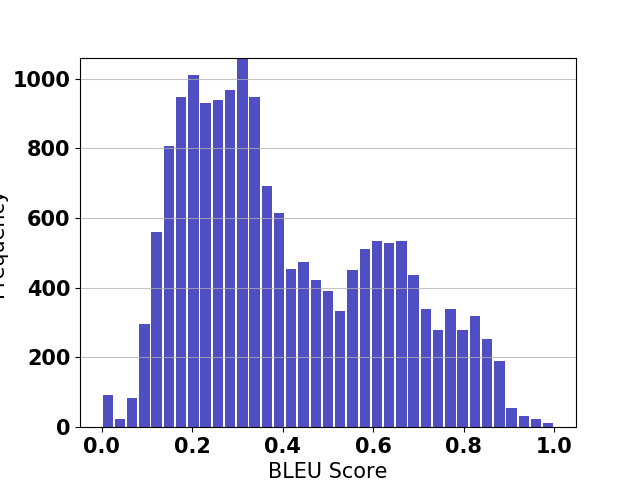}
    \caption{Overall}\label{fig:bleuoverall}
    \end{subfigure}
\caption{BLEU scores calculated between each worker-produced CA and the original argument (claim and premise). The results indicate that workers used keywords directly from the argument.}
\label{fig:similarity}
\end{figure*}
In order to determine how annotators created their CAs, we calculate the BLEU~\cite{papineni2002bleu} score of each CA and the argument (e.g.,\ premise/claim). The distribution in Figure~\ref{fig:similarity} indicates that workers copied keywords directly from the original argument in some cases.

\paragraph{\textbf{How similar are the CAs across annotators?}}
One design decision when building Riposte!\@ was that with more annotators, we could collect a wide variety of diverse CAs for a single-argument regardless of the fallacy type. We first calculate the similarity of the CAs across annotators for a single argument. We tokenize the corpus using spaCy\footnote{\url{https://spacy.io/}} and remove stop words and punctuation. We then calculate the average Jaccard similarity score for all combinations of CAs per unique argument and average over all arguments. The results (see Table~\ref{tab:diversity}) indicate that the CAs are diverse. 



\section{Experiment}


\subsection{Experimental design}
\label{design}
In Section~\ref{corpus_eval}, we observed that workers copied keywords from the argument when creating a CA.
Based on this observation, we experiment with different input settings to the model to better understand which parts of the argument annotators used to create their argument (e.g.,\ topic (T) only, premise (P) only, claim (C) only, and so forth). We cast the task of automatic constructive feedback as a generation task and experiment with such settings.

Since both new and existing essay topics can be used and introduced by educators, we consider two possible settings: i) \textbf{in-domain} (i.e.\ topics are shared between splits) and ii) \textbf{out-of-domain} (i.e.\ topics are not shared).

For our generation model, we use gold fallacy type information.\footnote{We built an LSTM-encoder multi-label classifier and the results of 4-way classification was 36.02\% F1 score, indicating more sophisticated features such as background knowledge and reasoning are necessary.} This allows us to understand how well the model can generate CAs when correct fallacy types are predicted.

\subsection{Data preparation}

We filter out all \textit{unsure} instances. We use majority vote for selecting CAs and their fallacy types. We split the data into 80\% train, 10\% test, and 10\% dev. In each setting, we ensure that no unique claim-premise pairs are shared across splits.

For each experiment, we tokenize using spaCy and lowercase all tokens. For \hg CAs, we replace the template with a special token (i.e.\ \textit{$<$hg$>$}). For all other CAs, we discard the original template and add a special token between slot-fillers.
This allows our baseline model to focus more on the content words found in the original argument.

\subsection{Baselines}
Based on our observations in Section~\ref{design}, we create a baseline for determining which parts of an argument annotators used to create CAs and how well a model can generate a CA.

\paragraph{\textbf{Simple Overlap (SO)}}
We calculate simple BLEU overlap for each setting against the CA as a baseline. In order to directly compare the results to our seq2seq baseline model, we calculate the BLEU scores for the preprocessed data from our seq2seq baseline model with unknown words.



\paragraph{\textbf{Seq2Seq}}

We preprocess and train our model using fairseq~\cite{ott2019fairseq}. We use pre-trained word embeddings (300-dimensional GloVe embeddings~\cite{pennington2014glove}) which are useful for generation tasks~\cite{qi2018}. We create two models (seq2seq-i and seq2seq-o) for in-domain and out-of-domain settings, respectively.\footnote{For seq2seq-i and seq2seq-o, we use the best hyperparameters from seq2seq-i (P+C) and seq2seq-o (P+C) across all settings, respectively.}

\begin{table}[t]
    \centering
    \begin{adjustbox}{max width=\linewidth}
    \begin{tabular}{l|r|r|r|r|r|r|r}
    \toprule
    \textbf{Baseline} & \textbf{T} & \textbf{C} & \textbf{P} & \textbf{T+P+C} & \textbf{T+C} & \textbf{T+P} & \textbf{P+C}\\
    \midrule
    \textbf{SO} & 3.98 & 6.37 & 15.59 & 13.56 & 10.69 & 13.76 & 18.16\\\midrule
    \textbf{seq2seq-i} & 12.28 & 12.31 & 5.96 & 14.54 & 12.63 & 13.37 & 16.57\\
    \textbf{seq2seq-o} & 1.31 & 1.05 & 1.49 & 4.78 & 1.60 & 1.53 & 5.53\\

    \bottomrule
    \end{tabular}
    \end{adjustbox}
    \caption{BLEU scores of our baselines using gold fallacy type for topic (T), premise (P), and claim (C).}
    \label{tab:params}
\end{table}





\subsection{Evaluation}

\begin{table}[t!]
    \centering
    \begin{adjustbox}{max width=\linewidth}
    \begin{tabular}{l|r|r|r|r}
         \toprule
         \textbf{Attribute} & \textbf{Scores (GO)} & \textbf{$\alpha$ (GO)} & \textbf{Scores (GE)} & \textbf{$\alpha$ (GE)}\\
         \midrule
    Strength & 2.3 & 0.20 & 1.98 & 0.20 \\
     Persuasiveness & 2.26 & 0.71 & 1.94 & 0.15 \\
     Relevance &  2.74 & 0.20 &  2.84 & 0.72 \\
         \bottomrule
    \end{tabular}
    \end{adjustbox}
    \caption{Mean scores and agreement (Krippendorff's $\alpha$) scores for gold (GO) and generated (GE) CAs.}
    \label{tab:resultsgenerated}
\end{table}

We evaluate the results of our baselines using BLEU (see Table~\ref{tab:params}). Our SO results indicate that workers mainly used the premise and claim when creating CAs. We observe that seq2seq-o's performance is low, indicating a simple model is not sufficient when unknown topics are introduced.

For evaluation, we would also like to compare the quality of gold CAs against generated CAs. We conduct an annotation study using AMT (3 workers per CA) and evaluate CA quality using 3 dimensions: \textit{Strength}, \textit{Persuasiveness}, and \textit{Relevance}.\footnote{We use \newcite{carlile2018give}'s guidelines and slightly modify for CAs. Please see the Appendix for our criteria.}
In total, we show 50 arguments and their gold/generated CAs, where each argument is annotated by 3 workers.\footnote{We use 50 generated CAs from seq2seq-i (P+C).} The results are shown in Table~\ref{tab:resultsgenerated}.\footnote{We convert from a 5 to 3-scale for score calculation.} We observed that workers found generated CAs more relevant, but the arguments were weaker and less persuasive.
Examples of the generated output for our best model (seq2seq-i P+C) are shown in Table~\ref{tab:seq2seqexamples}.




\begin{table}[t!]
\centering
\tiny
\begin{tabular}{p{2.7cm}|p{1.8cm}|p{2cm}}
\toprule
\textbf{Source} & \textbf{Reference} & \textbf{Hypothesis}\\
\midrule
home - schoolers should play for high school teams because all children should be able to participate in sports . \btq{} & all children are to play in sports \btq{} even home - schoolers will be playing sports .
 & all children should be able to participate in sports \btq{} home - schoolers should play for high school teams .
\\\hline
the u.s . should lift sanctions with cuba because the embargo hurts our own economy . \rh{} &  the u.s .\rh{}  the embargo .& us sanctions \rh{} our own economy .\\
\bottomrule
\end{tabular}
\caption{Examples of output from seq2seq-i (P+C).}
\label{tab:seq2seqexamples}
\end{table}

\section{Conclusion and future work}

In this work, we construct \textbf{Riposte!}, a large corpus of 18,887 crowdworker-produced CAs. Our analysis on Riposte!\ reveals that non-expert crowdworkers can produce reasonably diverse CAs. 
We cast automatic constructive feedback as a text generation task and create a baseline model. 

In our future work, we will explore injecting background knowledge and reasoning into our model to generate CAs for unknown topics and provide detailed information to students about how to improve their original argumentation.

\bibliographystyle{acl_natbib}
\bibliography{emnlp-ijcnlp-2019}

\appendix

\section{Annotation Interface and Guidelines}
\begin{figure}[t!]
\begin{center}
\includegraphics[clip, width=.5\textwidth,height=10cm]{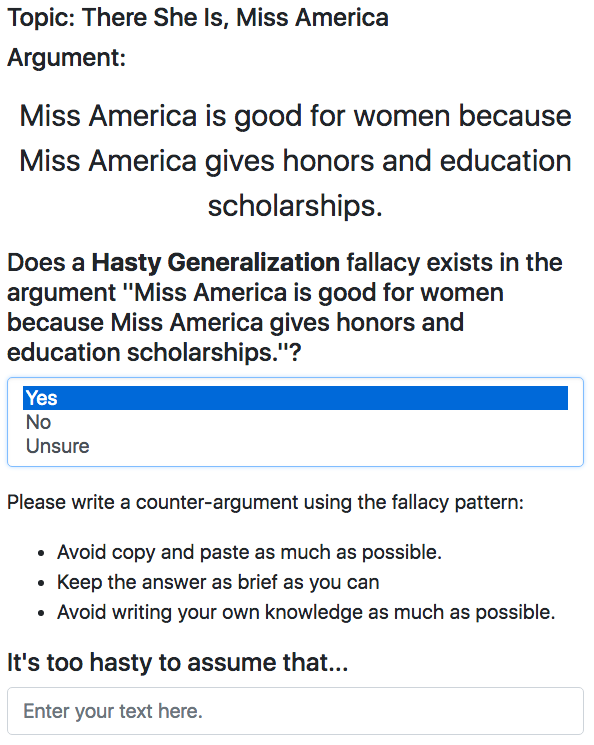}
\caption{Interface shown to crowdworkers for our \textit{hasty generalization} full-fledged experiment.}
\label{fig:interface_hasty}
\end{center}
\end{figure}

\begin{figure}[t]
\begin{center}
\includegraphics[clip, width=.5\textwidth]{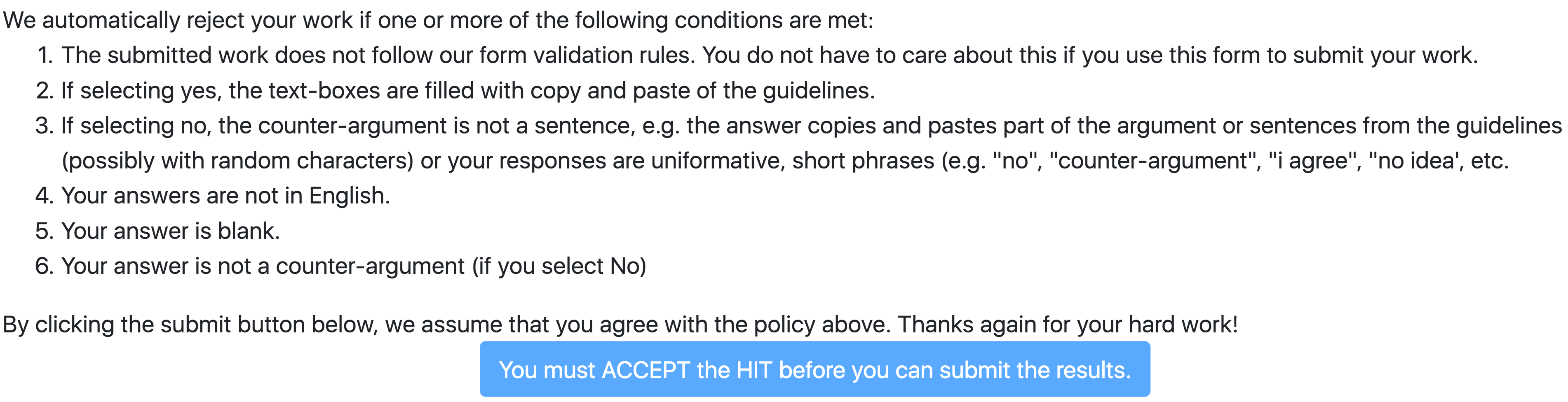}
\caption{Conditions for rejecting worker's responses shown to workers for \btq, \qc, and \rh experiments.}
\label{fig:conditions}
\end{center}
\end{figure}

\begin{figure}[t]
\begin{center}
\includegraphics[clip, width=.5\textwidth]{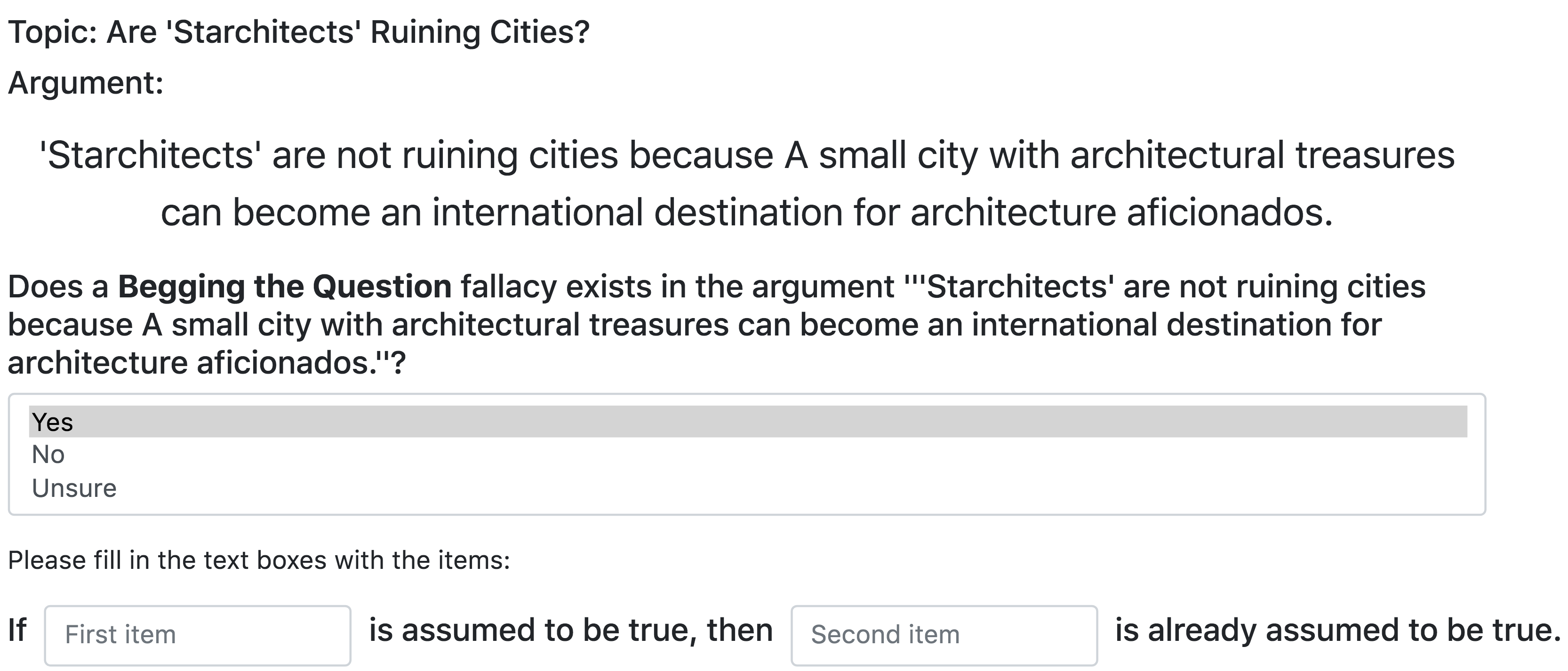}
\caption{Interface for \btq.}
\label{fig:begging}
\end{center}
\end{figure}

We show the annotation interface used in our full-fledged crowdsourcing experiment in Figure~\ref{fig:interface_hasty}. The conditions shown to workers for 3 fallacy types are shown in Figure~\ref{fig:conditions}. The interface for \btq{} is shown in Figure~\ref{fig:begging}.

\begin{figure}[t!]
\begin{center}
\includegraphics[clip, width=.5\textwidth]{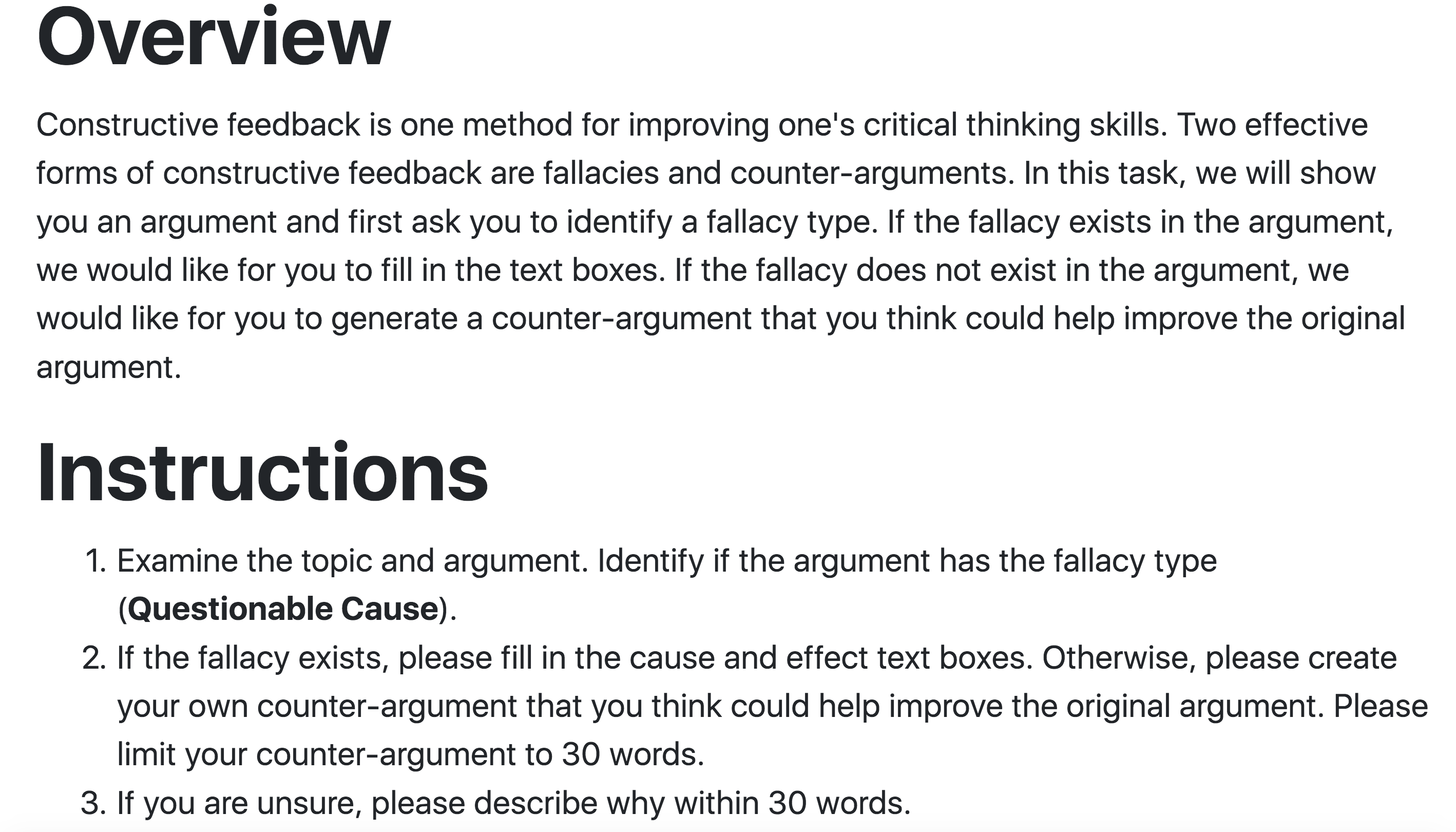}
\caption{Guidelines shown to crowdworkers.}
\label{fig:guidelines}
\end{center}
\end{figure}

 The guidelines shown to workers is shown in Figure~\ref{fig:guidelines}.

\begin{figure}[t]
\begin{center}
\includegraphics[clip, width=.5\textwidth]{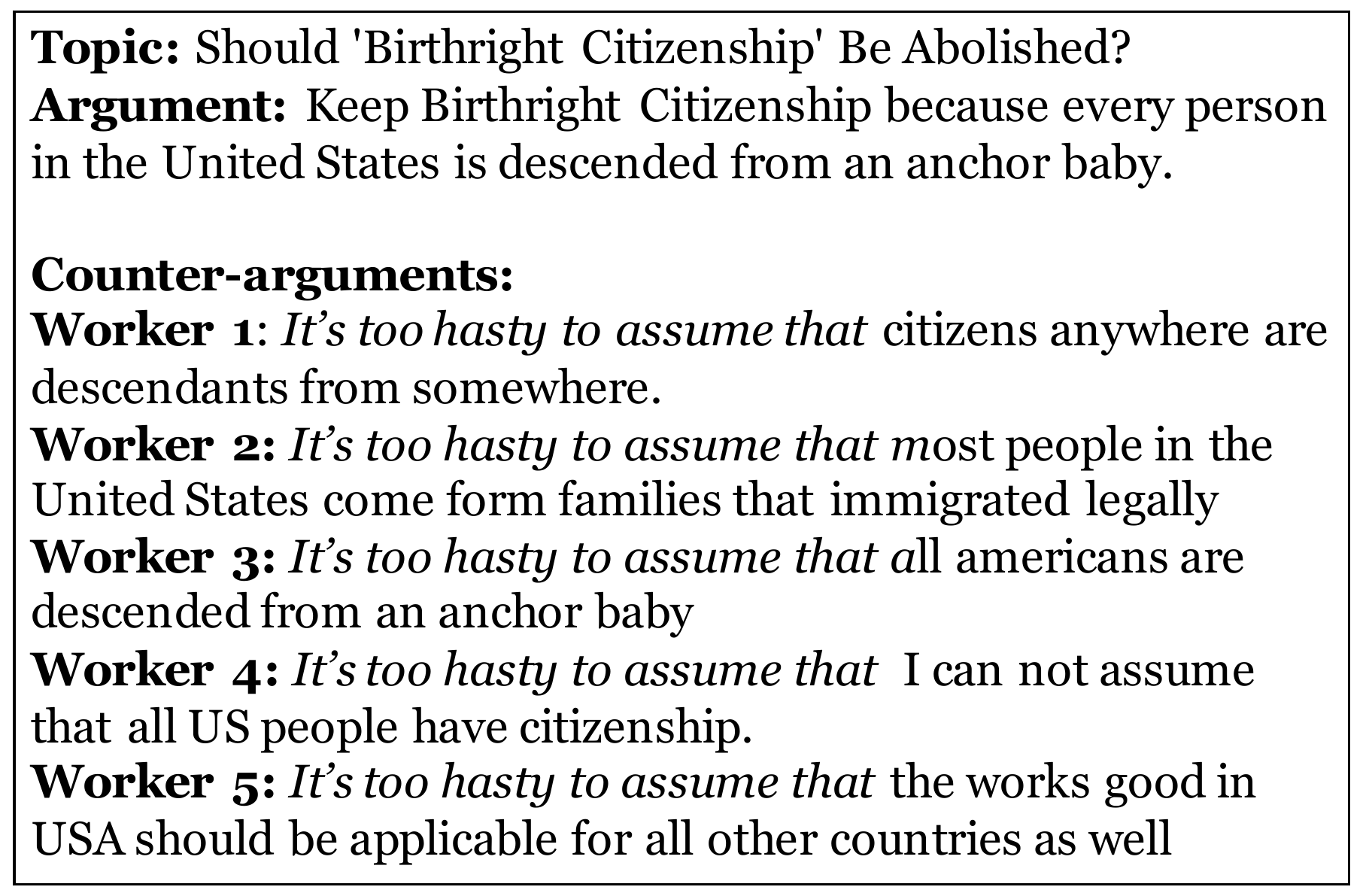}
\caption{CAs produced for a single argument (\textit{hasty generalization}) with perfect annotator agreement. All 5 workers agreed the fallacy existed.}
\label{fig:diversity}
\end{center}
\end{figure}

\section{Crowdsourcing settings}

For our full-fledged experiment, we use the following settings: workers were required to have a number of Human Intelligence Tasks (HITs) approved to be greater than or equal to 100 and a HIT Approval Rate greater than or equal to 96\%. For each HIT, workers were rewarded with \$0.20 (in the case of \textit{hasty generalization}, workers were rewarded with \$0.10).  An example of the guidelines for one fallacy type (e.g., questionable cause) are shown in Figure~\ref{fig:guidelines}. For each of our experiments below, the settings are as follows. If workers selected \textit{no} or \textit{unsure}, they were required to provide a CA or reason, respectively. We inform workers that their work will be rejected if one or more of the following conditions is met. The CA is i) blank, ii) not a sentence, iii) a direct copy-paste of the original argument in the text box or copy-paste of the guidelines, or iv) not written in English. We manually reject responses that fall under this criteria.

\section{Model Hyperparameters}

\begin{table}[ht]
    \centering
    \begin{tabular}{l|r}
    \toprule
    \textbf{hyperparameter} & \textbf{values}\\
    \midrule
    dropout  & {0.1, 0.2, 0.3, 0.4, 0.5} \\
    encoder/decoder layers & {1,2,3}\\
    hidden layers & {128, 256, 512, 1024}\\
    learning rate & {0.1, 0.01, 0.001}\\
    optimizers & {adam, sgd}\\
    \bottomrule
    \end{tabular}
    \caption{Hyperparameters used in our experiments for seq2seq-i (P+C) and seq2seq-o (P+C).}
    \label{tab:hyparams}
\end{table}

For seq2seq-i (P+C) and seq2seq-o (P+C), we experiment with the hyperparameters shown in Table~\ref{tab:hyparams}.
The best hyperparameters for our experiment are as follows. For seq2seq-i, we use the following settings. The dropout is set to 0.4. We use SGD as an optimizer with a learning rate of 0.01. The number of encoder/decoder layers is set to 1, and the encoder/decoder hidden size is 256.

For seq2seq-o, we use the following settings. The dropout is set to 0.2. We use SGD as an optimizer with a learning rate of 0.01. The number of encoder/decoder layers is set to 1, and the encoder/decoder hidden size is 256.

\begin{table*}[t]
    \centering
    \begin{tabular}{l|p{12.5cm}}
         \toprule
         \textbf{Attribute} & \textbf{Description (Strong)} \\
         \midrule
    \textbf{Relevant} & Anyone can see how the counter-argument attacks the argument. The relationship between the two components is either explicit or extremely easy to infer. The relationship is thoroughly explained in the text because the two components contain the same words or exhibit coreference.\\
    \textbf{Persuasive} & A very strong, clear counter-argument. It would persuade most readers and is devoid of errors that might detract from its strength or make it difficult to understand.\\
    \textbf{Strength} & A very strong counter-argument with no fallacies. Not much can be improved in order to attack the argument better.\\
         \bottomrule
    \end{tabular}
    \caption{Guidelines for annotating the quality of the CAs in our corpus, where the description is shown for the highest score (5). Each dimension has a score of 1-5. Annotators are only shown the criteria for the highest and lowest score only.}
    \label{tab:criteria}
\end{table*}

\section{Annotation Criteria and Examples}
The guidelines shown to crowdworkers when annotating the quality of CAs are shown in Table~\ref{tab:criteria}. We show the description for strong dimensions (i.e., score of 5).

Examples of CAs for one argument are shown in Figure~\ref{fig:diversity}.

\end{document}